\documentclass[conference]{IEEEtran}
\IEEEoverridecommandlockouts
\usepackage{cite}
\usepackage{amsmath,amssymb,amsfonts}
\usepackage{algorithmic}
\usepackage{graphicx}
\usepackage{textcomp}
\usepackage{xcolor}
\def\BibTeX{{\rm B\kern-.05em{\sc i\kern-.025em b}\kern-.08em
    T\kern-.1667em\lower.7ex\hbox{E}\kern-.125emX}}
\begin{document}

\title{Learn to Jump: Adaptive Random Walks for Long-Range Propagation through Graph Hierarchies\\

}
\newcommand{\abs}[1]{\left|#1\right|}
\newcommand{\oo}[1]{\mathcal{O}\left(#1\right)}
\newcommand{\CITE}{{\color{red} [CITE]}}
\newcommand{\E}{\mathbb{E}}

\author{\IEEEauthorblockN{Joël Mathys}
\IEEEauthorblockA{
\textit{ETH Zurich}\\
jmathys@ethz.ch}
\and
\IEEEauthorblockN{ Federico Errica}
\IEEEauthorblockA{
\textit{NEC Italia}\\
federico.errica@neclab.eu}

}

\maketitle

\begin{abstract}
Message-passing architectures struggle to sufficiently model long-range dependencies in node and graph prediction tasks. We propose a novel approach exploiting hierarchical graph structures and adaptive random walks to address this challenge. Our method introduces learnable transition probabilities that decide whether the walk should prefer the original graph or travel across hierarchical shortcuts. On a synthetic long-range task, we demonstrate that our approach can exceed the theoretical bound that constrains traditional approaches operating solely on the original topology. Specifically, walks that prefer the hierarchy achieve the same performance as longer walks on the original graph. These preliminary findings open a promising direction for efficiently processing large graphs while effectively capturing long-range dependencies.
\end{abstract}


\section{Introduction}
\label{sec:introduction}

The field of graph machine learning studies how to leverage the information contained in a graph -- an abstract representations for a set of interconnected entities -- to solve node and graph prediction tasks in a data-driven manner \cite{bacciu_gentle_2020}. As of today, the message-passing paradigm \cite{scarselli_graph_2009, micheli_neural_2009} remains the most prominent approach to graph processing, where nodes exchange messages between each other and iteratively increase their contextual information as a result. In this paper, we are interested in solving tasks where it is presumably required to consider \textit{long-range} interactions between distant nodes. 

Capturing long-range dependencies in graphs poses a significant challenge because of how information flows through the network topology.
In standard message-passing architectures, nodes synchronously exchange messages at each iteration, causing an exponential growth in receptive fields as the number of iterations increases. This inevitably forces models to compress vast amounts of information into fixed-size node embeddings, which can lead to significant information loss -- a problem known as oversquashing \cite{alon_on_2021}. For tasks requiring long-range dependencies, traditional  message-passing architectures rely on many iterations (or graph convolution layers), which can worsen the oversquashing problem.
%One limitation of most message-passing architectures is that all nodes synchronously exchange messages at every iteration, which can result in an exponential growth of nodes' receptive fields as the number of iterations increases. This means that a great deal of information needs to be compressed into a fixed-size node embedding, causing information loss: this problem is called \textit{oversquashing} \cite{alon_on_2021}, and it generally happens regardless of the number of message-passing iterations used. On tasks where long-range dependencies are important, classical message-passing architectures rely on a high number of iterations (also called ``graph convolutional layers") to capture such dependencies, with a consequent emergence of oversquashing. 
In this sense, oversquashing is \textit{one} of the potential barriers to solve long-range propagation tasks on graphs.

Two promising approaches have emerged to address long-range challenges. First, hierarchical methods introduce virtual structures above the original graph, reducing topological distances between distant nodes.
%A popular approach to address long-range tasks is to introduce a hierarchical structure on top of the graph {\color{red} [CITE]}, which can greatly reduce the topological distance between two nodes.
Current message-passing approaches based on hierarchies mostly require ad-hoc definitions of graph convolutional layers for the hierarchy \cite{hierarchical_rampasek,zhu_hignn_2022,dong_megraph_2023}, but this does not necessarily mitigate oversquashing, even if one reduces the number of message-passing iterations, as the coarse-graining process into hierarchical nodes may cause the same process of information loss. In contrast, the second approach, namely random walks (RWs), has shown promising results to mitigate these effects while capturing long-range interactions. 
%Because RWs select a restricted set of nodes to compute a source node's representation, there is no exponential growth of information to be squashed into each node, so oversquashing should be mostly mitigated. 
Because RWs have access to the original node embeddings while walking, there should be less squashing of information into the individual nodes. Moreover, they proved being beneficial %as a feature-engineering approach that 
to augment node features with structural embeddings \cite{dwivedi2022graph}. However, due to the uninformed walking process, it is unclear how likely and well they can capture long range phenomenons unless they walk the entire graph.

In light of these facts, the contribution of this work is to combine the use of a hierarchical strucure with \textbf{adaptive} RW transition probabilities to approach long-range problems defined on graphs. This way, the model can decide whether it makes sense to walk the hierarchy or the original graph.
% while mitigating the oversquashing effect typical of synchronous message-passing architectures. 
%We show, on a synthetic dataset where we can bound the expected error based on the size of the graph, that learning to walk the hierarchy provides a sensible advantage compared to random walks and/or the absence of the hierarchical topology.
We validate our approach on a synthetic dataset with provable bounds on expected performance, 
where we demonstrate that our adaptive hierarchical walks substantially outperform standard random walk approaches operating on either the original or hierarchical topology alone. Specifically, we show that our method can exceed the theoretical bound that constrains approaches operating solely on the original graph structure, achieving with shorter walks what would require significantly longer traversals in conventional approaches.

\section{Related Work}

Walk-based approaches provide an effective alternative to graph representation learning. Early work such as DeepWalk \cite{deepwalk} introduced treating such walks as sentences and nodes as words.
Node2Vec\cite{node2vec} extended this by biasing the walking strategy in favour of balanced exploration, while AgentNet \cite{martinkus2023agentbased} aimed to fully learn the walk policy. CRAWL \cite{tonshoff2023walking} incorporated additional topology information directly into walk processing with sequence models. Recent success on applying state space models for graph learning, as demonstrated by GraphMamba \cite{graphmamba24}, has led to renewed interest in combining these techniques with random walks \cite{chen2025learning, kim2025revisiting,cho_walker}. However, current approaches remain limited to uninformed walks on the original graph topology when capturing long-range dependencies.

Hierarchical approaches offer valuable perspectives on multi-level graph modeling. METIS \cite{metis} established fundamental partitioning techniques, while recent work on Hierarchical Support Graphs \cite{vonessen2024levelmessagepassinghierarchicalsupport} and hierarchical learning \cite{hierarchical_rampasek} have shown how coarsened structures can improve the performance of message passing, often utilizing pooling methods \cite{pooling_survey} and set-based representations \cite{deepset}. While these approaches reduce distances between nodes in the graph, they may not be enough to overcome the challenges associated with message passing such as oversquashing or may even introduce new information bottlenecks for it.

\section{Method}

We propose to combine hierarchical graph structures with learnable random walks to address long-range interactions in graphs. 
The key insight of our approach is that these two paradigms offer complementary advantages. Hierarchical structures provide efficient shortcuts between distant regions of the graph, while random walks avoid compression of information stored in the nodes compared to message-passing.  
By introducing a coarser virtual structure above the original graph, we create alternative paths between distant nodes, effectively reducing their topological distance. The model can learn to navigate between the detailed information in the original graph and the hierarchy as needed for the task. This allows us to capture long-range patterns using significantly shorter walks than would be required when operating solely on the original topology.

Our approach consists of three main components: a hierarchical graph structure that provides multiple levels of abstraction, an adaptive random walk mechanism with learnable transition probabilities, and a walk processing module that embeds walk sequences into node representations.
Figure 1 illustrates how our approach facilitates efficient long-range information propagation compared to standard random walks on the original graph.

\textbf{Hierarchical Graph}
Given an input graph $G = (V, E)$, we construct a hierarchical graph $G_H = (V_H, E_H)$ where $V_H = V \cup V_1 \cup \ldots \cup V_L$ and $E_H = E \cup E_1 \cup \ldots \cup E_L \cup E_{parent}$. Here, $V$ represents the original graph nodes (level 0), while $V_l$ represents nodes at hierarchical level $l$. Each new level of the hierarchy represents a coarser variant of the level below it, preserving topological structure through $E_l$. Moreover, each node connects to exactly one parent in the upper level through $E_{parent}$. This process continues until the top level contains only a single node. We use the METIS partitioning algorithm \cite{metis} to build this hierarchical structure, which has been shown effective for creating appropriate hierarchical support graphs \cite{vonessen2024levelmessagepassinghierarchicalsupport}. We set the refinement factor to $\frac{1}{2}$, meaning that each new level has approximately half the nodes of the previous level. This ensures that the overall size of the hierarchical graph $G_H$ remains linear in the size of the original graph $\abs{V_H} = \oo{|V|}$, while the number of levels $L$ is logarithmic $\oo{\log |V|}$.

The features $x_v^l$ of the new virtual nodes are initialized using the mean of all descendants $u\in V_0$ that are original vertices $x^0_u$. Additionally, all edges are augmented with their respective type and direction (up, down, horizontal).
Then, we learn hierarchical node embeddings through a bottom-up formulation using DeepSet pooling\cite{deepset}. For each level $l$ in the hierarchy, we aggregate information from nodes at level $l-1$ to inform representations at level $l$.  For each node $v \in V_l$, we first transform the embeddings of its children $\mathcal{C}(v) \subset V_{l-1}$ , aggregate them through summation, and combine this aggregated information with $v$'s current representation:
\begin{equation}
\mathbf{h}^l_v = \phi_{\text{out}}([\mathbf{x}^l_v \mid\mid \sum_{u \in \mathcal{C}(v)} \phi_{\text{in}}(\mathbf{h}^{l-1}_u)])
\end{equation}
Where $h^0_v = x^0_v$, $\mid\mid$ denotes concatenation and the $\phi$ are implemented as MLPs. This process is repeated for each hierarchical level, enabling information to flow from the original graph to the coarsest level of abstraction.\\
%The resulting embeddings capture multi-scale structural information, with higher levels representing increasingly broader contexts.

\begin{figure}[ht]
\centerline{
\includegraphics[width=0.5\textwidth]{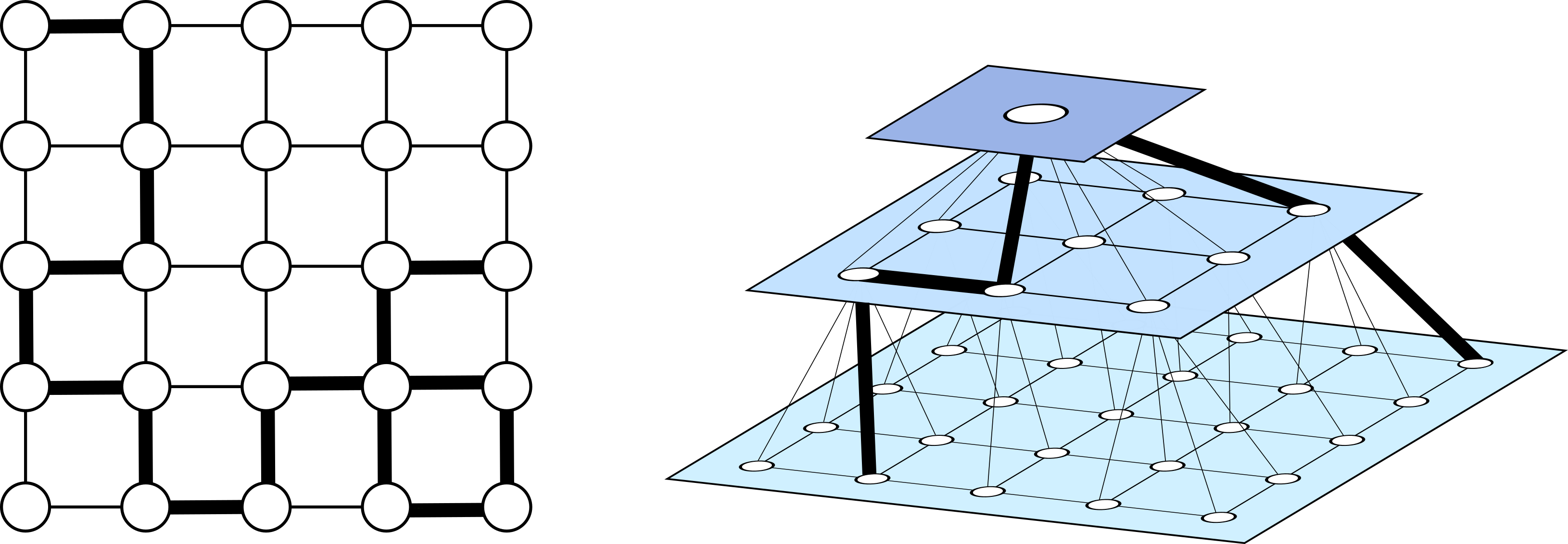}
}
\caption{Starting from the original graph topology, we create a hierarchical coarsened graph structure on top of it. On the original graph topology, capturing long range interactions with random walks might require very long walks, whereas the same can be achieved with adaptive random walks through the hierarchy. The hierarchical connections enable efficient 'jumps' between distant regions of the graph.}
\label{fig:story}
\end{figure}

\textbf{Adaptive Walks}
At the core of our approach is the adaptive random walk mechanism that learns to navigate the original and/or hierarchical graph structures. 
For each node $v \in V_0$, we sample $k$ walks of length $L$ starting from $v$. Unlike standard random walks that use uniform transition probabilities, we use a learnable transition function $q$ implemented as a MLP, which determines the probability of moving from node $u \in V_H$ to node $v \in V_H$:
\begin{equation}
\Pr(v|u) = \frac{\exp(q(h_u, h_v, e_{uv}))}{\sum_{w \in \mathcal{N}(u)} \exp(q(h_u, h_w, e_{uw}))}
\end{equation}
Where $\mathcal{N}(u)$ denotes the neighbors of $u$, including both the original and hierarchical edges. We use the Gumbel-Softmax \cite{jang2017categorical} for differentiable discrete sampling, enabling an end-to-end training procedure. Importantly, to prevent the walker from immediately returning to its previous position, we mask out the previous node during sampling.
This adaptive mechanism is crucial for efficiently capturing long-range dependencies, as it allows our model to learn and trade off navigating the original graph for local details and to leverage hierarchical shortcuts to reach distant regions.

\textbf{Walk Aggregation} Each sampled walk $v_0, v_1, ..., v_{L-1}$ results in a sequence of node embeddings, which we process using a Mamba sequence model \cite{gu2024mamba}. Mamba has been shown to be effective for processing RW on graphs \cite{chen2025learning}, to derive a sequence of embeddings per walk: 
\begin{equation}
z_w = \Phi([h_{v_0}, h_{v_1}, \ldots, h_{v_{L-1}}]) = [s_{v_0}, s_{v_1}, ..., s_{v_{L-1}}]    
\end{equation}
Where $\Phi$ is a Mamba model. Finally, we aggregate these embeddings to the origin nodes of the walks for all $v \in V_0$.
\begin{equation}
{o_v} = \mathbf{h}_v + \phi \left( \frac{1}{kL}\sum_{w: v_0 = v}s_{v_i}\right)
\end{equation}
The resulting node embedding combines the original node representation with information from $k$ adaptive walks and can then be used for the corresponding downstream task.

\section{Experiments}

We evaluate our method on the PrefixSum task \cite{mathys2024floodechonetalgorithmically}, a controllable benchmark in which we can isolate and measure a model's ability to capture long-range interactions. For our experiments, we construct undirected line graphs with $n=16$ nodes, where each node has a binary feature $x_i \in {0,1}$ sampled uniformly and its position. The task for node $i$ is to predict $y_i = (\sum_{j=0}^{i} x_j) \bmod 2$, which requires integrating information from all preceding nodes. This creates dependencies of varying lengths throughout the graph, with later nodes requiring longer-range dependencies. What makes this task particularly interesting is its provable theoretical bound: any model with walk length (or convolution depth) $L \leq n$, the expected accuracy over the data distribution $A$ is bounded by:
\begin{equation}
    \E[A] \leq 1 - \frac{\frac{n-L}{n}}{2} = \frac{1}{2} + \frac{L}{2n}
\end{equation}
\begin{figure}[ht]
\centerline{
\includegraphics[width=0.5\textwidth]{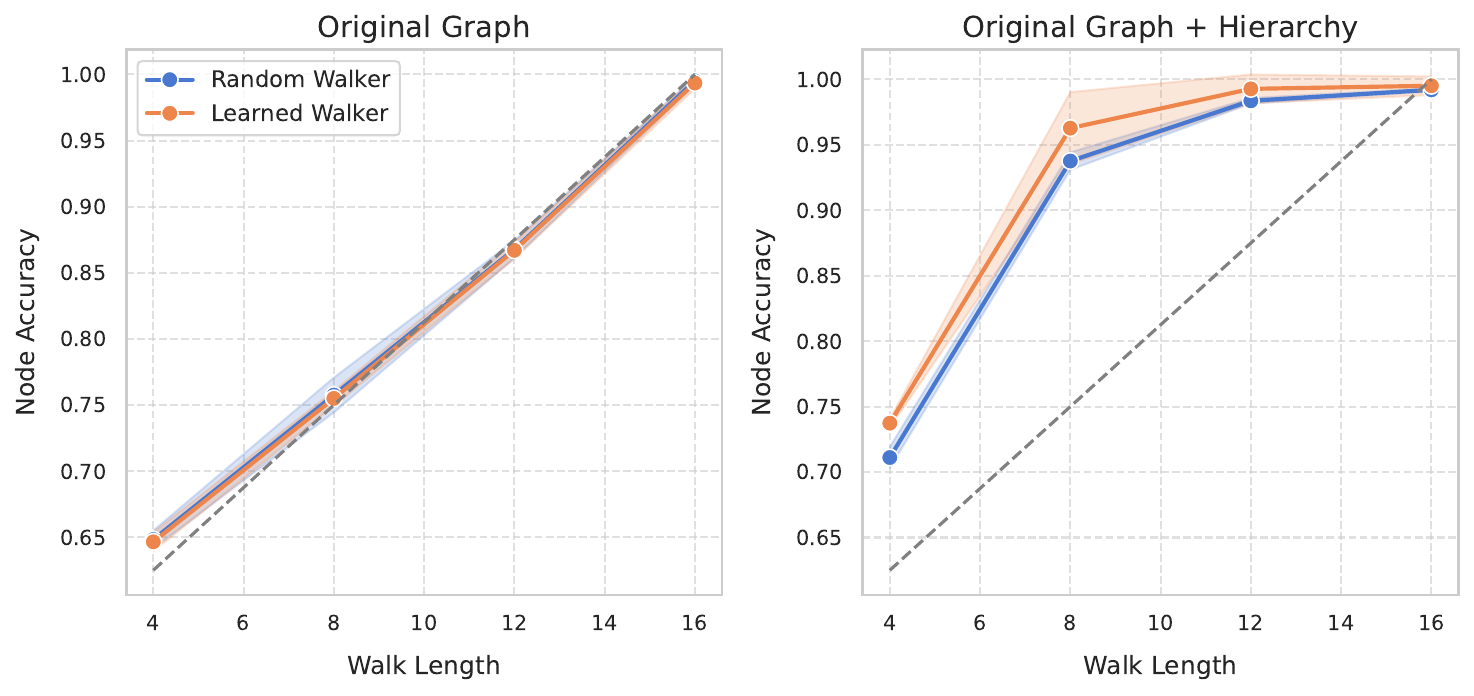}
}
\caption{Restricted to the graph, models cannot go beyond the expected accuracy if $L\!\leq\! n$. However, with our hierarchical approach (right), walks can gather information from much farther away, achieving performance that would require much longer walks on the original graph}
\label{fig:prefix}
\end{figure}
As $x_0$ can only be part of the receptive field of $(n-L)$ nodes, meaning the other nodes cannot deterministically determine their correct sum. Importantly, this bound applies to any model that operates on the original topology. 

Our dataset consists of 1000 randomly generated instances, split 80/10/10 for training/validation/testing. We train all models using AdamW for 200 epochs following a cosine learning rate schedule with an initial warmup of 5 epochs, and we set $k=10$. We report all node-level prediction accuracies averaged over 5 independent runs.
We evaluate the performance of our approach using both learned and random transitions as well as the role of the hierarchical graph construction. This setup allows us to isolate the effects of both the hierarchical structure and the adaptive walker, demonstrating their contributions to long-range dependency modeling.

\section{Results}

\begin{figure*}[ht]
\centerline{
\includegraphics[width=\textwidth]{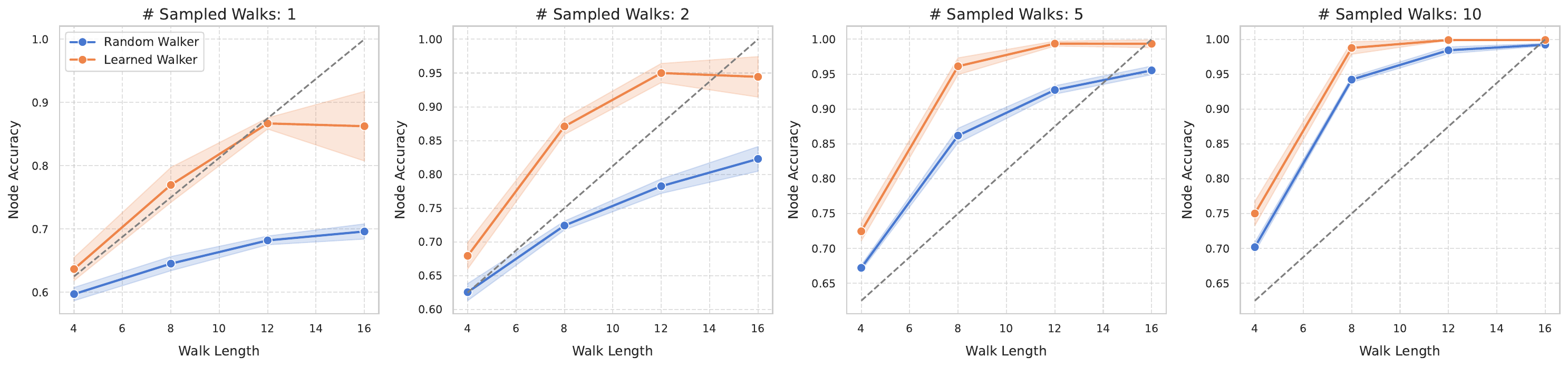}
}
\caption{When we decrease the number of sampled walks per node $k$ during inference, we observe the increasing advantage of learned hierarchical walks. While both of our variants perform better than the derived accuracy bound for classical approaches, the learned variant can do so sooner, using fewer walks. Note how even short learned walks match or outperform much longer random walks.}
\label{fig:prefix_nr_walks}
\end{figure*}

We first evaluate the methods on the original graph topology, illustrated in Figure~\ref{fig:prefix} (left). These results confirm that models operating solely on the original graph topology cannot exceed the theoretical accuracy bound of $0.5 + L/2n$, regardless of model architecture. As a consequence, all standard random walks methods fall at or below this bound regardless of their specific architecture. In contrast, Figure~\ref{fig:prefix} (right) shows that our method incorporating hierarchical walks can clearly go beyond this theoretical bound. By leveraging the hierarchical structure, our approach can achieve the same performance with walks of length $L$ that would require longer walks of length $>L$ on the original graph.
This demonstrates the benefit of the hierarchical structure, which fundamentally changes what is achievable within limited walk lengths.

Figure~\ref{fig:prefix_nr_walks} provides additional insights into the learned models, illustrating the performance depending on the number of sampled walks per node \textit{during inference}. As we decrease the number $k$ of walks per node, the advantage of learned walks over random walks becomes more substantial. Note that both variants make use of non-backtracking walks. Moreover, both variants of our method outperform the theoretical bound for classical approaches, however, the learned variant can efficiently achieve higher accuracy with fewer samples. Notably, these learned walks still match or exceed the performance of longer walks which are sampled randomly. This underlines the viability and effectiveness of adaptively navigating the hierarchical structure in a learned manner.

Summarizing, these results show that adaptive walks on a hierarchical structure allow for more effective propagation of information across long distances in the graph. The ability of our method to exceed the theoretical bound shows the benefit of incorporating coarsened structure. Therefore, these results open up promising directions of learning long range interactions using shorter walks of sublinear length, \textit{which could be especially interesting for very large graphs}.

\section{Conclusion}

In this work, we introduced a novel approach combining hierarchical graph structures with adaptive random walks to address the challenge of capturing long-range dependencies on graphs. Our key contribution lies in demonstrating that such hierarchical walks can overcome the limitations that constrain models operating solely on the original graph topology. Through experiments on the PrefixSum task, we showed that our method achieves the same performance  with walks of length $L$ that would require much longer walks on the original graph.
The results reveal two important findings: First, creating coarsened structures through a hierarchy changes what performance is achievable within given walk lengths. Second, learning adaptive transitions can significantly improve efficiency, enabling our model to maintain higher performance with fewer sampled walks compared to random traversal strategies. Moreover, the short learned walks achieve the same performance as longer random walks, even on the hierarchy.
These findings open up promising directions for capturing long-range interactions, particularly for applications to very large graphs, where  using shorter, more efficient walks could yield substantial advantages.

\bibliographystyle{IEEEtran}  % This uses the IEEE transaction citation style
\bibliography{refs}  % This will import citations from refs.bib

\begin{thebibliography}{10}
\providecommand{\url}[1]{#1}
\csname url@samestyle\endcsname
\providecommand{\newblock}{\relax}
\providecommand{\bibinfo}[2]{#2}
\providecommand{\BIBentrySTDinterwordspacing}{\spaceskip=0pt\relax}
\providecommand{\BIBentryALTinterwordstretchfactor}{4}
\providecommand{\BIBentryALTinterwordspacing}{\spaceskip=\fontdimen2\font plus
\BIBentryALTinterwordstretchfactor\fontdimen3\font minus \fontdimen4\font\relax}
\providecommand{\BIBforeignlanguage}[2]{{%
\expandafter\ifx\csname l@#1\endcsname\relax
\typeout{** WARNING: IEEEtran.bst: No hyphenation pattern has been}%
\typeout{** loaded for the language `#1'. Using the pattern for}%
\typeout{** the default language instead.}%
\else
\language=\csname l@#1\endcsname
\fi
#2}}
\providecommand{\BIBdecl}{\relax}
\BIBdecl

\bibitem{bacciu_gentle_2020}
D.~Bacciu, F.~Errica, A.~Micheli, and M.~Podda, ``A gentle introduction to deep learning for graphs,'' \emph{Neural Networks}, vol. 129, pp. 203--221, 9 2020.

\bibitem{scarselli_graph_2009}
F.~Scarselli, M.~Gori, A.~C. Tsoi, M.~Hagenbuchner, and G.~Monfardini, ``The graph neural network model,'' \emph{IEEE Transactions on Neural Networks}, vol.~20, no.~1, pp. 61--80, 2009.

\bibitem{micheli_neural_2009}
A.~Micheli, ``Neural network for graphs: {A} contextual constructive approach,'' \emph{IEEE Transactions on Neural Networks}, vol.~20, no.~3, pp. 498--511, 2009.

\bibitem{alon_on_2021}
U.~Alon and E.~Yahav, ``On the bottleneck of graph neural networks and its practical implications,'' in \emph{9th International Conference on Learning Representations (ICLR)}, 2021.

\bibitem{hierarchical_rampasek}
L.~Rampášek and G.~Wolf, ``Hierarchical graph neural nets can capture long-range interactions,'' in \emph{2021 IEEE 31st International Workshop on Machine Learning for Signal Processing (MLSP)}, 2021, pp. 1--6.

\bibitem{zhu_hignn_2022}
W.~Zhu, Y.~Zhang, D.~Zhao, J.~Xu, and L.~Wang, ``Hignn: Hierarchical informative graph neural networks for molecular property prediction equipped with feature-wise attention,'' \emph{arXiv}, 2022.

\bibitem{dong_megraph_2023}
H.~Dong, J.~Xu, Y.~Yang, R.~Zhao, S.~Wu, C.~Yuan, X.~Li, C.~J. Maddison, and L.~Han, ``Megraph: capturing long-range interactions by alternating local and hierarchical aggregation on multi-scaled graph hierarchy,'' in \emph{Proceedings of the 37th {Conference} on {Neural} {Information} {Processing} {Systems} ({NeurIPS})}, 2023.

\bibitem{dwivedi2022graph}
V.~P. Dwivedi, A.~T. Luu, T.~Laurent, Y.~Bengio, and X.~Bresson, ``Graph neural networks with learnable structural and positional representations,'' in \emph{10th International Conference on Learning Representations}, 2022.

\bibitem{deepwalk}
\BIBentryALTinterwordspacing
B.~Perozzi, R.~Al-Rfou, and S.~Skiena, ``Deepwalk: online learning of social representations,'' in \emph{Proceedings of the 20th ACM SIGKDD International Conference on Knowledge Discovery and Data Mining}, ser. KDD '14.\hskip 1em plus 0.5em minus 0.4em\relax New York, NY, USA: Association for Computing Machinery, 2014, p. 701–710. [Online]. Available: \url{https://doi.org/10.1145/2623330.2623732}
\BIBentrySTDinterwordspacing

\bibitem{node2vec}
\BIBentryALTinterwordspacing
A.~Grover and J.~Leskovec, ``node2vec: Scalable feature learning for networks,'' in \emph{Proceedings of the 22nd ACM SIGKDD International Conference on Knowledge Discovery and Data Mining}, ser. KDD '16.\hskip 1em plus 0.5em minus 0.4em\relax New York, NY, USA: Association for Computing Machinery, 2016, p. 855–864. [Online]. Available: \url{https://doi.org/10.1145/2939672.2939754}
\BIBentrySTDinterwordspacing

\bibitem{martinkus2023agentbased}
\BIBentryALTinterwordspacing
K.~Martinkus, P.~A. Papp, B.~Schesch, and R.~Wattenhofer, ``Agent-based graph neural networks,'' in \emph{The Eleventh International Conference on Learning Representations}, 2023. [Online]. Available: \url{https://openreview.net/forum?id=8WTAh0tj2jC}
\BIBentrySTDinterwordspacing

\bibitem{tonshoff2023walking}
\BIBentryALTinterwordspacing
J.~T{\"o}nshoff, M.~Ritzert, H.~Wolf, and M.~Grohe, ``Walking out of the weisfeiler leman hierarchy: Graph learning beyond message passing,'' \emph{Transactions on Machine Learning Research}, 2023. [Online]. Available: \url{https://openreview.net/forum?id=vgXnEyeWVY}
\BIBentrySTDinterwordspacing

\bibitem{graphmamba24}
\BIBentryALTinterwordspacing
A.~Behrouz and F.~Hashemi, ``Graph mamba: Towards learning on graphs with state space models,'' in \emph{Proceedings of the 30th ACM SIGKDD Conference on Knowledge Discovery and Data Mining}, ser. KDD '24.\hskip 1em plus 0.5em minus 0.4em\relax New York, NY, USA: Association for Computing Machinery, 2024, p. 119–130. [Online]. Available: \url{https://doi.org/10.1145/3637528.3672044}
\BIBentrySTDinterwordspacing

\bibitem{chen2025learning}
\BIBentryALTinterwordspacing
D.~Chen, T.~H. Schulz, and K.~Borgwardt, ``Learning long range dependencies on graphs via random walks,'' in \emph{The Thirteenth International Conference on Learning Representations}, 2025. [Online]. Available: \url{https://openreview.net/forum?id=kJ5H7oGT2M}
\BIBentrySTDinterwordspacing

\bibitem{kim2025revisiting}
\BIBentryALTinterwordspacing
J.~Kim, O.~Zaghen, A.~Suleymanzade, Y.~Ryou, and S.~Hong, ``Revisiting random walks for learning on graphs,'' in \emph{The Thirteenth International Conference on Learning Representations}, 2025. [Online]. Available: \url{https://openreview.net/forum?id=SG1R2H3fa1}
\BIBentrySTDinterwordspacing

\bibitem{cho_walker}
Y.~Wang and K.~Cho, ``Non-convolutional graph neural networks.'' in \emph{Advances in Neural Information Processing Systems}, A.~Globerson, L.~Mackey, D.~Belgrave, A.~Fan, U.~Paquet, J.~Tomczak, and C.~Zhang, Eds., vol.~37.\hskip 1em plus 0.5em minus 0.4em\relax Curran Associates, Inc., 2024, pp. 32\,705--32\,730.

\bibitem{metis}
\BIBentryALTinterwordspacing
G.~Karypis and V.~Kumar, ``A fast and high quality multilevel scheme for partitioning irregular graphs,'' \emph{SIAM Journal on Scientific Computing}, vol.~20, no.~1, pp. 359--392, 1998. [Online]. Available: \url{https://doi.org/10.1137/S1064827595287997}
\BIBentrySTDinterwordspacing

\bibitem{vonessen2024levelmessagepassinghierarchicalsupport}
\BIBentryALTinterwordspacing
C.~Vonessen, F.~Grötschla, and R.~Wattenhofer, ``Next level message-passing with hierarchical support graphs,'' 2024. [Online]. Available: \url{https://arxiv.org/abs/2406.15852}
\BIBentrySTDinterwordspacing

\bibitem{pooling_survey}
\BIBentryALTinterwordspacing
C.~Liu, Y.~Zhan, J.~Wu, C.~Li, B.~Du, W.~Hu, T.~Liu, and D.~Tao, ``Graph pooling for graph neural networks: progress, challenges, and opportunities,'' in \emph{Proceedings of the Thirty-Second International Joint Conference on Artificial Intelligence}, ser. IJCAI '23, 2023. [Online]. Available: \url{https://doi.org/10.24963/ijcai.2023/752}
\BIBentrySTDinterwordspacing

\bibitem{deepset}
M.~Zaheer, S.~Kottur, S.~Ravanbhakhsh, B.~P\'{o}czos, R.~Salakhutdinov, and A.~J. Smola, ``Deep sets,'' in \emph{Proceedings of the 31st International Conference on Neural Information Processing Systems}, ser. NIPS'17.\hskip 1em plus 0.5em minus 0.4em\relax Red Hook, NY, USA: Curran Associates Inc., 2017, p. 3394–3404.

\bibitem{jang2017categorical}
\BIBentryALTinterwordspacing
E.~Jang, S.~Gu, and B.~Poole, ``Categorical reparameterization with gumbel-softmax,'' in \emph{International Conference on Learning Representations}, 2017. [Online]. Available: \url{https://openreview.net/forum?id=rkE3y85ee}
\BIBentrySTDinterwordspacing

\bibitem{gu2024mamba}
\BIBentryALTinterwordspacing
A.~Gu and T.~Dao, ``Mamba: Linear-time sequence modeling with selective state spaces,'' in \emph{First Conference on Language Modeling}, 2024. [Online]. Available: \url{https://openreview.net/forum?id=tEYskw1VY2}
\BIBentrySTDinterwordspacing

\bibitem{mathys2024floodechonetalgorithmically}
\BIBentryALTinterwordspacing
J.~Mathys, F.~Grötschla, K.~V. Nadimpalli, and R.~Wattenhofer, ``Flood and echo net: Algorithmically aligned gnns that generalize,'' 2024. [Online]. Available: \url{https://arxiv.org/abs/2310.06970}
\BIBentrySTDinterwordspacing

\end{thebibliography}

\end{document}